\documentclass[letterpaper, 10 pt, conference]{style/ieeeconf}

\IEEEoverridecommandlockouts
\usepackage{graphics} 
\usepackage{epsfig} 
\usepackage{mathptmx} 
\usepackage{times} 
\usepackage{amsmath} 
\usepackage{amssymb}  
\usepackage{siunitx} 
\usepackage[inkscapelatex=false]{svg}
\usepackage{subcaption}
\usepackage[hidelinks]{hyperref}

\usepackage{subcaption} 
\usepackage{pdfpages} 

\usepackage{algorithm}
\usepackage{algpseudocode}

\usepackage{booktabs}

\usepackage{authblk}

\usepackage{calc}
\usepackage{array}
\usepackage{tabulary}
\renewcommand{\baselinestretch}{0.96}
\DeclareMathAlphabet{\mathcal}{OMS}{cmsy}{m}{n}

\usepackage{mathtools}

\title{\LARGE \bf
Sampling Strategies for Robust Universal Quadrupedal Locomotion Policies
}
\author[1]{David Rytz\thanks{
Corresponding author: rytz@robots.ox.ac.uk. Video can be found \color{blue}{\href{https://drive.google.com/file/d/1WOcTAHUVt3fSx6DaRGvJj6BABNcPOQ40/view?usp=sharing}{here}}.
}
}
\author[1]{Kim Tien Ly}
\author[1]{Ioannis Havoutis}
\affil[1]{Dynamic Robot Systems, Oxford Robotics Institute, University of Oxford}
\begin{document}
%
\maketitle
\begin{abstract}
This work focuses on sampling strategies of configuration variations for generating robust universal locomotion policies for quadrupedal robots. We investigate the effects of sampling physical robot parameters and joint proportional-derivative gains to enable training a single reinforcement learning policy that generalizes to multiple parameter configurations. Three fundamental joint gain sampling strategies are compared: parameter sampling with (1) linear and polynomial function mappings of mass-to-gains, (2) performance-based adaptive filtering, and (3) uniform random sampling. We improve the robustness of the policy by biasing the configurations using nominal priors and reference models. All training was conducted using the RaiSim simulation environment, tested in simulation on a range of diverse quadrupeds, and zero-shot deployed onto hardware using the ANYmal quadruped robot. Compared to multiple baseline implementations, our results demonstrate the need for significant joint controller gains randomization for robust closing of the sim-to-real gap. 
\end{abstract}

\section{Introduction}

\label{sec:introduction}
The increasing maturity of quadrupedal robotic systems has led to a wide range of applications, including industrial inspection, surveillance, research, and search and rescue operations. Advances in numerical optimization~\cite{mastalliCrocoddylEfficientVersatile2020, grandiaPerceptiveLocomotionNonlinear2023a}, Reinforcement Learning (RL)~\cite{  zhuangRobotParkourLearning2023, kimHighspeedControlNavigation2025}, and combined~\cite{jeneltenDTCDeepTracking2024a} control approaches made essential contributions to the development of quadrupedal locomotion and navigation in complex environments with increased agility. 

Most existing control designs are tailored to specific robots, optimized for a single platform, and require extensive retraining and parameter tuning whenever the robot's morphology or dynamics change. These platform-specific approaches pose a fundamental scalability challenge: developing a controller for each new quadruped requires notable computational resources, laborious manual parameter tuning and reward engineering, as well as hardware expertise. As the diversity of commercial quadrupeds continues to grow -- from lightweight such as the A1 (\SI{12}{kg},~\cite{unitreeA12023}) to larger ANYmal (\SI{50}{kg},~\cite{anyboticsANYmal2025a}) -- such individualized design architectures become increasingly impractical.

In this paper, we identify three central challenges for achieving \textit{universal} quadrupedal locomotion. First, quadrupeds exhibit substantial variation in mass distribution, joint configurations, actuator properties, and kinematic structures, which hinders the design of policies that generalize across platforms. Second, these morphological differences directly affect locomotion dynamics, requiring different gait patterns and control strategies, and thereby complicating controller design, hardware transfer, and performance evaluation. Third, a suitable robot parameter generation approach is needed to support efficient and stable training across a wide range of configurations, despite the absence of known optimal proportional-derivative (PD) gain schedules for previously unseen robots.

\begin{figure}
    \centering
    \includegraphics[width=0.95\columnwidth]{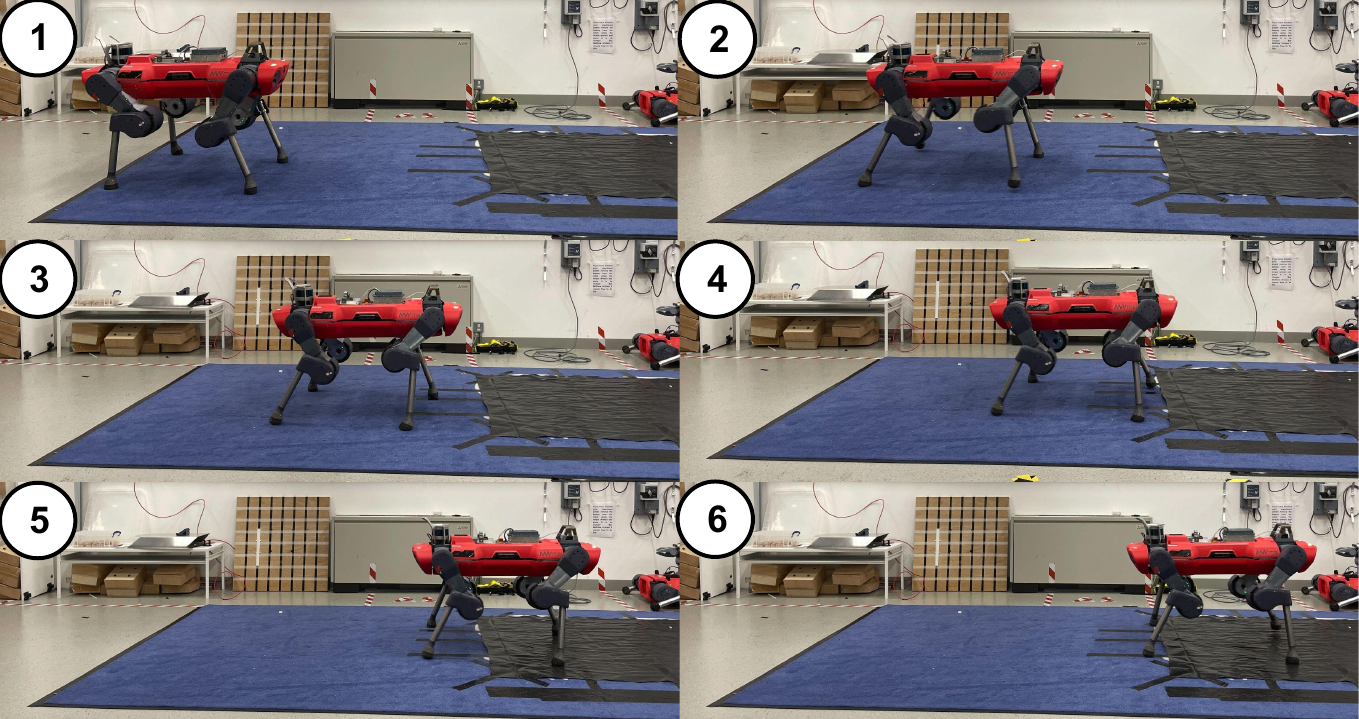}
    \caption{Example run of ANYmal hardware used. Stand still at frames 1 and 6.}
    \vspace{-0.5cm}
    \label{fig:results:hardware}
\end{figure}

To address these challenges, we investigate two approaches to sampling robot configurations and compare three strategies for PD gain tuning. As demonstrated in the supplementary video, strong sim-to-sim performance does not guarantee reliable hardware behavior. We validate our methods in simulation on a range of quadrupeds and in hardware experiments on the ANYmal platform (Fig.~\ref{fig:results:hardware}), using models that were not encountered during training.

Our contributions include: (1) a novel sampling method of morphologies for a \textit{universal} RL locomotion policy with a wide range of kinematics and dynamics randomization, (2) improving on state-of-the-art sampling methods for joint PD gain tuning and comparison on how these parameters affect locomotion performance, and (3) successful zero-shot transfer to various quadrupeds in simulation (ranging from the A1, with \SI{12}{\kg}, to ANYmal, with \SI{50}{kg}) and on hardware, including the ANYmal robot as shown in Fig.~\ref{fig:results:hardware}.

\section{Related Work}
\label{sec:related-work}

\subsection{Reinforcement Learning for Legged Locomotion}
Deep RL has emerged as a powerful paradigm for developing robust and agile controllers for legged robots, with early success including policies trained for dynamic and agile quadrupedal skills~\cite{LearningAgileDynamica} and enabling robust rough terrain locomotion in indoor and outdoor settings~\cite{mikiLearningRobustPerceptive2022, choiLearningQuadrupedalLocomotion2023}. More recent works include highly dynamic parkour maneuvers~\cite {kimHighspeedControlNavigation2025}, wheeled-legged locomotion~\cite{leeLearningRobustAutonomous2024}, as well as whole-body multi-legged control~\cite{ maLearningCoordinatedBadminton2025}. 

Standard RL pipelines often rely on extensive reward shaping, long training times, and remain sensitive to morphology-specific variations to achieve natural and smooth locomotion motions. Motion imitation, based on animal motion capture~\cite{ pengLearningAgileRobotic2020}, partial demonstrations~\cite{liLearningAgileSkills2023}, adversarial motion priors~\cite{escontrelaAdversarialMotionPriors2022}, or optimal control-based reference motions generated by a trajectory optimizer~\cite{fuchiokaOPTMimicImitationOptimized2023} have become a common strategy to acquire mammal-like smooth gaits with reduced reward engineering.

A major challenge with RL policy is sim-to-real transfer, mostly addressed by using one of the two strategies: (1) applying domain randomization during training to increase robustness~\cite{pengSimtoRealTransferRobotic2018a}, or (2) increasing simulation fidelity via accurate system modelling~\cite{xieLearningLocomotionSkills2020} and identification ~\cite{LearningAgileDynamica}. While the latter approach aligns dynamics in simulation with reality, domain randomization provides general robustness by exposing the learned policy to a range of varying parameters such as ground and motor friction, latency, and other robot or environment parameters. 

\subsection{Universal Locomotion Control}

Recent research has increasingly focused on \textit{universal} locomotion controllers that generalize across different robot morphologies, actuation, and sensing setups. One major line of work incorporates morphology into the policy architecture, often using Graph Neural Networks (GNN) \cite{huangOnePolicyControl2020} or transformers \cite{guptaMetaMorphLearningUniversal2022a, trabuccoAnyMorphLearningTransferable2022} to represent robot structure. While these methods demonstrate promising transfer capabilities, most have been validated primarily in simulation, with limited real-world results on light-weight robots typically below \SI{30}{\kilo\gram}. Meta-RL methods provide another perspective, with techniques such as Model-Agnostic Meta-Learning (MAML)~\cite{finnModelAgnosticMetaLearningFast2017} enabling policies to adapt rapidly to new morphologies~\cite{belmonte-baezaMetaReinforcementLearning2022a}. These approaches aim for few-shot adaptation, meanwhile, our work achieves zero-shot hardware deployment.

A critical factor in enabling universal locomotion is the sampling strategy used to expose policies to diverse morphologies during training. Prior controller architectures that were deployed on hardware differ significantly in this respect:
\begin{itemize}
    \item \textbf{GenLoco}~\cite{fengGenLocoGeneralizedLocomotion2022} utilizes imitation learning based on motion capture data combined with a large state-action history to train a phase-variable locomotion controller. They use a fixed set of reference robot models and uniformly randomized morphology parameters such as masses, inertias, control latency, actuator properties, and link sizes, mapping the resulting total robot mass to joint PD gains via a linear function with uniform scaling. 
    \item \textbf{ManyQuadrupeds}~\cite{shafieeManyQuadrupedsLearningSingle2023} relied on Central Pattern Generators (GPG) for controls but handled morphology diversity and robot models similarly to GenLoco. Using CPGs necessitates this control architecture to rely on robot-specific inverse kinematics. No domain randomization is applied.


    \item \textbf{MorAL}~\cite{luoMorALLearningMorphologically2024} embeds robot-specific information using a morphology encoder trained from a single reference model, combined with extensive domain randomization of masses, link dimensions, motor gains, and
    actuator dynamics.
    

    \item \textbf{URMA}~\cite{bohlingerOnePolicyRun2025} employs a transformer-based policy across multiple legged morphologies and applies domain randomization around robot-specific nominal parameters. Its relatively narrow sampling ranges limit robustness despite strong simulation and hardware performance.

    \item \textbf{PAL}~\cite{rytzReferenceFreePlatform2025} extends MorAL to a broader set of robot models and demonstrates zero-shot deployment on the \SI{50}{\kilo\gram} ANYmal platform by uniformly sampling robot parameters, including PD gains, from wide predefined ranges.
\end{itemize}

These approaches have been shown to function effectively on small to medium-sized robots (below \SI{30}{\kg}), though often with reduced smoothness and stability as robot size increases. PAL demonstrated that broader model diversity is critical for scaling locomotion policies to heavier hardware. Building on this observation, we introduce a novel morphology parameter sampling method based on particle filter techniques, inspired by their use in terrain generation~\cite{leeLearningQuadrupedalLocomotion2020}. 

\section{Preliminaries - Reinforcement Learning}
\label{sec:preliminaries}


We formulate the locomotion problem as a Markov Decision Process in an RL setting. At each discrete time step $t$ the agent observes the state $\mathbf{s}$ and takes action $\mathbf{a}$ transitioning to state $\mathbf{s_{i+1}}$ with transition probability $\mathcal{P}(\mathbf{s}_i, \mathbf{a}_i, \mathbf{s}_{i+1})$. The agent receives a scalar reward $\mathrm{r}_i = \mathcal{R}(\mathbf{s}_i, \mathbf{a}_i, \mathbf{s}_{i+1})$ with -$\mathcal{R}$ being the reward function. Starting from an initial state $s_0$, this process continues until a terminal condition is reached, such as a finite time horizon or success/failure. The objective that the agent is maximizing is the expected cumulative discounted reward:
\begin{equation}
    J\left(\pi\right)\doteq\underset{\mathcal{T}\sim\pi_\theta}{\text{\textup{E}}}\left[\sum_{t=0}^{\infty}{{\gamma}^{t}R\left(\mathbf{s}_{t},\mathbf{a}_{t},\mathbf{s}_{t+1}\right)}\right]\text{,}
    \label{eq:discounted_reward_ex}
\end{equation}
where $\gamma\in\left[0,1\right)$ is the discount factor and $\mathcal{T}$ denotes the trajectory induced by policy $\pi$ and $\mathrm{\pi}: \mathbf{s} \rightarrow ~a$ is a parametrized mapping from states to actions. In this work, $\pi$ represents a \textit{universal} locomotion controller, mapping observations to actions such that quadrupeds with diverse morphological properties achieve consistent and stable behavior.

\section{Methodology}
\label{sec:methodology}

The following section details the proposed pipeline illustrated in Fig.~\ref{fig:methodology:control_framework}, which includes the architectures of the dynamics encoding network (shown in red), the control policy (shown in blue), and the generation of morphology and control parameters for training (shown in green).

\subsection{Universal Locomotion Controller}\label{ss:UniversalLocomotionController}
The control framework used was built following the asymmetric actor critic architecture first presented in MorAL~\cite{luoMorALLearningMorphologically2024}. 


\begin{figure*}[t]
    \centering

    \begin{subfigure}[c]{0.6\textwidth}
        \centering
        \vspace{0pt}
        \includegraphics[width=\linewidth]{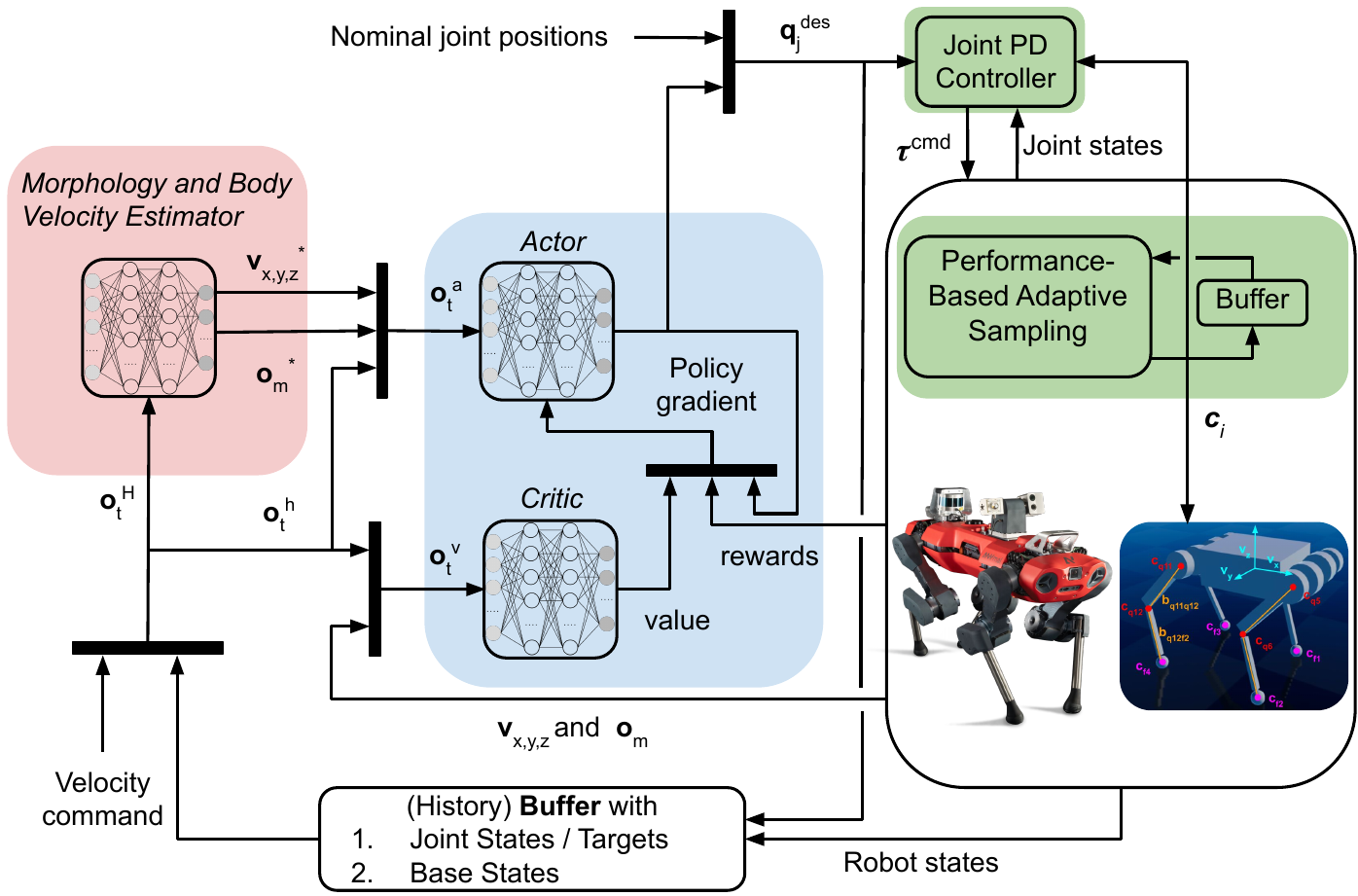}
        \caption{}   \label{fig:methodology:control_framework_a}
    \end{subfigure}
    \hfill
    \begin{subfigure}[c]{0.37\textwidth}
        \vspace{0pt}
        \textbf{Performance-Based Adaptive Sampling:}
        \vspace{2pt}

        {\footnotesize
        \begin{algorithmic}[1]
        \State \textbf{Initialize:} $\text{SR} \gets 0.1$, $\mathbf{c}_i \gets \mathbf{c}_i^{\text{nom}}$, Buffer $\mathcal{B} \gets \emptyset$
        \Loop \ every $N_{\text{traj}}$ rollouts
            \State \textit{// 1. Evaluate Performance \& Update Curriculum}
            \For{each robot $i$}
                \State Compute $Tr_i^{\text{lin}}$ and $Tr_i^{\text{zero}}$
                \State Compute $w_i^k$ \hfill $\triangleright$ Eq.~(\ref{eq:w_k_i_particleUp})
            \EndFor
            \State SR $\gets$ Update(SR) \hfill $\triangleright$ Eq.~(\ref{eq:sr})
            \State $\text{SR} \gets \text{clamp}(\text{SR}, 0.1, 1.0)$

            \State \textit{// 2. Resample and Validate Configurations}
            \State $\mathcal{S}_{new} \gets \emptyset,\; \mathcal{S}_{v} \gets \emptyset$
            \While {$|\mathcal{S}_{new}| < n_{\text{SR}}$} \hfill \textit{(Scaling)}
                \State $\mathbf{c}_{candidate} \sim \mathcal{U}(\mathbf{c}_i^{nom}, \text{SR})$
                \If{RaiSim valid}
                    \State Add $\mathbf{c}_{candidate}$ to $\mathcal{S}_{new}, \mathcal{S}_{v}$
                \EndIf
            \EndWhile

            \While {$|\mathcal{S}_{new}| < n_{\text{SR}} + n_w$} \hfill \textit{(Particle Filter)}
                \State $\mathbf{c}_{candidate} \gets$ SIR + NN walk
                \If{RaiSim valid}
                    \State Add $\mathbf{c}_{candidate}$ to $\mathcal{S}_{new}, \mathcal{S}_{v}$
                \EndIf
            \EndWhile

            \State \textbf{Update buffer} $\mathcal{B} \gets \mathcal{B} \cup \mathcal{S}_{v}$
            \State \textit{// 3. Diversity Preservation (Replay)}
            \State $\mathcal{S}_{new} \gets \mathcal{S}_{new} \cup \{\text{sample } n_r \text{ from } \mathcal{B}\}$
            \State \textbf{Update training set with $\mathcal{S}_{new}$}
        \EndLoop
        \end{algorithmic}
        }
        \vspace{-0.3cm}
        \caption{}   \label{fig:methodology:control_framework_b}
    \end{subfigure}

    \caption{
    (a): Motivated by~\cite{luoMorALLearningMorphologically2024, rytzReferenceFreePlatform2025}, the control framework stores base and joint states in a buffer and feeds them to the morphology and base velocity estimator (red). The actor produces desired joint positions $\mathbf{q}^{des}$ (blue) executed through a PD controller for quadruped locomotion. Candidate morphologies and actuator parameters (green) are generated using the Performance-Based Adaptive Sampling procedure described in (b), with further details in Section~\ref{ss:robot_generation}.}
    \label{fig:methodology:control_framework}
\end{figure*}

The input states for each module (red for estimation and blue for actor and critic), as per Fig.~\ref{fig:methodology:control_framework_a} in our control architecture, are:
\begin{enumerate}
    \item \textbf{Estimator:} By not using imitation learning, the architecture necessitates estimating base linear velocity $\mathbf{v}_\text{x,y,z}^*$. This network is extended to also observe robot-specific morphology parameters $\mathbf{o}_m^*$ consisting of base body centre of mass $\textbf{com}_{\text{base},{x,y,z}}\in\mathbb{R}^3$, body masses $\mathbf{m}_\text{base, hip, thigh, shank}\in\mathbb{R}^4$, joint link offsets $\mathbf{c}_{q_1^\text{xyz}, q_2^\text{xyz}, q_3^\text{xyz}}\in\mathbb{R}^9$ and foot offset $c_{f,z}\in\mathbb{R}^1$. 
    The network receives the robot state history $\mathbf{o}_t^H=\langle\mathbf{o}^h_{t=0},\mathbf{o}_{t=1}^h,\ldots, \mathbf{o}_{t=5}^h\rangle$ as input. We define
    \begin{equation}
        \mathbf{o}_{t=0}^h = \langle \mathbf{e}_z^B, \mathbf{\omega}_B, \Delta\mathbf{q_t}, \mathbf{\dot{q}_t}, \mathbf{q_{t-1}^\text{des}}, \mathbf{v}_{x,y,\omega}^\text{des} \rangle,
    \end{equation}
    consisting of robot base orientation $\mathbf{e}_z^B\in\mathbb{R}^3$ and angular velocity $\mathbf{\omega}_B\in\mathbb{R}^3$, joint position offset from nominal position $\Delta\mathbf{q_t}\in\mathbb{R}^{12}$, joint velocity $\mathbf{\dot{q}_t}\in\mathbb{R}^{12}$, previous joint action $\mathbf{q_{t-1}^
    \text{des}}\in\mathbb{R}^{12}$, and desired velocity command $\mathbf{v}_{x,y,\omega}^\text{des}\in\mathbb{R}^3$ as current robot state for $t=0$.
    
    \item \textbf{Actor:} The actor generates the desired joint position $\mathbf{q}_j^\text{des}\in\mathbb{R}^{12}$ action by receiving the input $\mathbf{o}_a$ comprised of estimated base linear velocity $\mathbf{v}_\text{x,y,z}^*$ and morphology parameters $\mathbf{o}_m^*$, the current robot state $\mathbf{o}_{t=0}^h$ and joint nominal position $\mathbf{q}^n\in\mathbb{R}^{12}$:
    \begin{equation}
        \mathbf{o}_t^a = \langle \mathbf{v}_\text{x,y,z}^*, \mathbf{o}_{t=0}^h, \mathbf{q}^n,  \mathbf{o}_m^* \rangle.
    \end{equation}
    The action is sampled from a Gaussian distribution with zero mean and standard deviation  $\sigma_a = 0.6$, added to the nominal joint configuration before passing it through the joint PD controller.
    \item \textbf{Critic:} The critic receives additional privileged information that would be difficult to infer during hardware deployment, including foot contact states $\mathbf{c}_t^{f, i}\in\mathbb{R}^4$, feet contact friction $\mathbf{\mu}^{f, i}\in\mathbb{R}^4$, feet height $\mathbf{h}_{t}^{f, i}$ and joint PD gains $\mathbf{k}_{PD}\in\mathbb{R}^{24}$. Instead of inputting the estimated morphology parameters, it processes the privileged parameters from the simulation environment, resulting in the input:
    \begin{equation}
        \mathbf{o}_t^v = \langle \mathbf{v}_\text{x,y,z}, \mathbf{o}_{t=0}^h, \mathbf{q}^n,  \mathbf{o}_m, \mathbf{c}_t^{f, i}, \mathbf{\mu}^{f, i}, \mathbf{h}_{t}^{f, i}, \mathbf{k}_{PD} \rangle.
    \end{equation}
\end{enumerate}

Compared to \cite{luoMorALLearningMorphologically2024, rytzReferenceFreePlatform2025}, we included the centre of mass of the base robot body into our architecture and removed the height map information, reducing training time and simplifying the evaluation of this work. 

\subsection{Reward Structure}
The reward is implemented following~\cite{rytzReferenceFreePlatform2025} with the weights used for the sum of its terms described in Table~\ref{tab:fp_reward_definitions}:

\begin{table}[ht!]
    \centering
    \footnotesize
    \setlength{\tabcolsep}{3pt}
    \begin{tabular}{|l|c||r|c|}
        \hline
        \textbf{Component} & \textbf{Weight} & \textbf{Component} & \textbf{Weight} \\
        \hline
        Base linear velocity & 3.0 & Joint position & -0.2 \\
        Base angular velocity & 1.5 & Joint velocity & $-3\!\times\!10^{-4}$ \\
        Base orientation & -5.0 & Joint acceleration & $-2\!\times\!10^{-7}$ \\
        Base height & -20.0 & Joint torque & $-3\!\times\!10^{-5}$ \\
        Base undesired motion & -0.5 & Joint action smoothness & -0.12 \\
        Foot slip & -0.2 & Joint action smoothness 2 & -0.05 \\
        Air time & -6.0 & & \\
        \hline
    \end{tabular}
        \caption{Reward term weights.}
        \label{tab:fp_reward_definitions}
            \vspace{-4mm}
\end{table}

All policies are trained with the same reward structure and weights, thus resulting in a similar locomotion gait pattern.

The task for this controller structure is to follow a velocity command of dimension $\mathbb{R}^3$ expressed in base frame as $\mathbf{c}^\text{des} = [\mathrm{v}_x^\text{des}\mathbf{e}_x^B \quad \mathrm{v}_y^\text{des}\mathbf{e}_y^B \quad
    \mathrm{\omega}_z^\text{des}\mathbf{e}_z^B]^T$. The command during training is uniformly sampled within the ranges $v_x^\text{max} = \pm \SI{1}{m/s}$, $v_y^\text{max} = \pm \SI{0.75}{m/s}$, and $\omega_z^\text{max} = \pm \SI{1.5}{rad/s}$ for durations of \SI{3}{s} to \SI{6}{s}.

\subsection{Joint Controller}
As depicted in Fig.~\ref{fig:methodology:control_framework_a} on the top in green, we use a Proportional-Derivative (PD) controller for each joint $j\in\left\{1,\ldots,12\right\}$  to regulate its desired position $\text{q}_j^\text{des}$ output from the actor. The controller outputs the demanded torque ${\tau}_j^\text{cmd}$ which is then applied to the actuators as
\begin{equation}
    \tau_j^{\text{cmd}} = K_p(\mathrm{q}^{\text{des}}_j-\mathrm{q}_j)
    - K_d\mathrm{\dot{q}}_j. 
    \label{eq:impedance_controller_simplified}
\end{equation}
with $K_p$ and $K_d$ being the proportional and derivative gains. During training, we randomize between the joint PD controller in the form of Equation~(\ref{eq:impedance_controller_simplified}) and the readily available Raisim physics simulator's built-in joint PD controller~\cite{raisimtechRaiSimV117Documentation2023}. As we could not guarantee joint torque limits to be enforced during training with the Raisim PD controller, we clip $\tau_j$ between the actuator torque limits $[-\tau^{max}_j, \tau^{max}_j]$. 

\subsection{Robot Configuration Generation and Randomization} 
\label{ss:robot_generation}
\subsubsection{Quadruped reference models}
Following~\cite{rytzReferenceFreePlatform2025}, we utilize four simplified reference models Unitree's A1~\cite{unitreeA12023} and Aliengo quadrupeds, as well as ANYbotics' ANYmal B~\cite{hutterANYmalHighlyMobile2016a} (an older and smaller version) and ANYmal C~\cite {ackermanevanANYboticsIntroducesSleek2019}. Their nominal values are used to resample a new set of parameters $\mathbf{c}_i$ to generate 40 robot configurations per robot model by randomizing their kinematic and dynamic properties.

A generic quadruped is modeled as a set of rigid bodies $b_{AB}$ with mass $m_{AB}\in\mathbb{R}^+$, connected by either fixed or revolute joints, where $A$ and $B$ denote parent and child joints, respectively. The base body $b_\text{base}$ is treated as a floating joint, and the relative placement of child joints is defined by $c_B\in\mathbb{R}^3$. Each foot is assigned a friction coefficient $\mu_f$ and a local frame offset $c_{f,z}$. Leg geometries are sampled from two distinct configurations: type A (knees pointing backward) or type X (knees pointing toward the base).

\subsubsection{Robot model parameter generation}\label{ss:robotModelGen}
Based on the \textit{reference quadrupeds}, we uniformly sample the morphology parameters similar to~\cite{rytzReferenceFreePlatform2025}, showing zero-shot deployment onto heavy-duty quadrupedal robots such as the ANYmal. The approach described for the estimator in Section~\ref{ss:UniversalLocomotionController} is extended by the base center of mass com$_\text{base,x}$ sampled uniformly $\mathcal{U}(-0.15,0.15)$ and com$_\text{base,y,z}$ as $\mathcal{U}(-0.1,0.1)$ added to $\mathbf{c}_i$ for all reference models. A configuration is admitted to the training set only if the robot remains collision-free in its nominal pose for \SI{2}{\second} using the RaiSim~\cite{raisimtechArticulatedSystemsRaiSim} joint controller.

The configuration sampling algorithm is further described by the pseudo-code in Fig.~\ref{fig:methodology:control_framework_b}.

We investigate the following strategies for sampling $n_p$ robot configurations with parameters $\mathbf{c}_i$ during training after $N_\text{traj}=\SI{40}{\second}$ simulated steps:

\textbf{Uniform randomization (baseline):} $n_u$ configurations are sampled from fixed uniform distributions. While providing broad coverage, this strategy lacks adaptation to training progress. This represents the current state-of-the-art, where the sampling range (SR) is typically fixed (e.g., $\text{SR}=1$ for GenLoco, ManyQuadrupeds, MorAL, and PAL, or to a more narrow range $\text{SR}=\text{small}$ for URMA). Different strategies for sampling joint PD gains under uniform randomization are discussed in detail in Section~\ref{ss:pdGainsRandomization}. 




\textbf{Adaptive curriculum sampling (ours):} We propose replacing fixed randomization with a performance-based adaptive curriculum that reshapes the configuration distribution during training. The mechanism combines two strategies: (1) adaptive clipping of uniform sampling ranges (SR) based on command-tracking performance, and (2) particle-filter-based resampling with local exploration.

Both mechanisms are driven by the same performance metrics and are applied jointly after every $N_{\text{traj}}$ rollouts.

\begin{itemize}
    \item \textbf{Performance-based scaling of sampling ranges:} We generate $n_\text{SR}$ configurations from the robot nominal parameters $\mathbf{c}_i^\text{nom}$ by clipping the SR of the \textbf{uniform randomization} to $\text{SR}\in\left(0.1,1\right)$. Training is initialized with $\text{SR}=0.1$ to ensure stable early learning, similar in spirit to URMA, but unlike URMA the SR is adapted online according to policy performance. We define the instantaneous linear and zero velocity thresholds inspired by the terrain-based traversability~\cite{leeLearningQuadrupedalLocomotion2020}:
    \begin{align}
        \nu^{\text{lin}}(t) &= 
        \begin{cases} 
            1 & \text{if } |\nu_{pr}^{\text{lin}}| > 0.3 \cdot \| [v_x^{\text{max}}, v_y^{\text{max}}] \| \\ 
            0 & \text{otherwise} 
        \end{cases} \\
        \nu^{\text{zero}}(t) &= 
        \begin{cases} 
            1 & \text{if } \|\mathbf{v}_{x,y,\omega}^{\text{cur}}\| < 0.2 \\ 
            0 & \text{otherwise} 
        \end{cases}
    \end{align}
    where $\nu^\text{lin}_{pr}$ is the scalar projection of the current base linear velocity onto the command in the robot base frame. $||\mathbf{v}_{x,y,\omega}^\text{cur}||$ is the norm of the current base velocity during a zero velocity command. The aggregate tracking performance $Tr_{i}^{\text{cmd}}$ for robot $i$ is the mean of these indicators over the active command duration $T_{\text{tot}}$ over all training environments. The ideal policy would achieve a value of 1 for the tasks $\text{cmd} \in \{\text{lin, zero}\}$, whereas a failing policy converges to 0. Computing the population mean, $Tr_{\text{mean}}^{\text{cmd}}$, we then update the SR after $N_\text{traj}$ as:
    \begin{equation}
        \text{SR} =
        \begin{cases}
        \text{SR} + 0.01, & \text{if } Tr_\text{mean}^\text{lin} > 0.65 \wedge Tr_\text{mean}^\text{zero} > 0.55, \\
        \text{SR} - 0.01, & \text{if } Tr_\text{mean}^\text{lin} < 0.55 \vee Tr_\text{mean}^\text{zero} < 0.40, \\
        \text{SR}, & \text{otherwise.}
        \end{cases}
        \label{eq:sr}
    \end{equation}
    
    \item \textbf{Particle filtering weighted resampling:} Simultaneously, we maintain a distribution over parameter sets and update it through a Sequential Importance Resampling (SIR) particle filter, inspired by~\cite{leeLearningQuadrupedalLocomotion2020}. The importance weight for a given $\mathbf{c}_i$ is used to update $n_{w}$ configurations using a random nearest neighbor walk and is defined as: 
    \begin{equation}\label{eq:w_k_i_particleUp}
        w^k_i = \frac{1}{2}\cdot \sum^{cmd}  Tr_i^\text{lin}\in[0.4,0.9] \wedge Tr_i^\text{zero}\in[0.4,0.9]. 
    \end{equation}
    The performance ranges are chosen so that only parameters yielding mid-to-upper-range performance are used, ensuring that training emphasizes robot parameter configurations $\mathbf{c}_i$ that are challenging but solvable.
    
    We additionally sample $n_{r}$ configurations from a replay buffer of previously sampled configurations to prevent degeneration of the parameter set. 
\end{itemize}
Joint nominal position configuration and joint torque limit parameters are sampled with SR$=1$ for all cases, enabling early adaptation to default standing poses and understanding of the resulting policies' torque limits.

\subsubsection{Joint PD Gains Randomization}\label{ss:pdGainsRandomization}
A central design choice in universal locomotion is the selection of robot joint PD gains $c_i^\text{PD}$. We compare four strategies:
\begin{itemize}
    \item \textbf{Mass-dependent mapping:} Similar to GenLoco~\cite{fengGenLocoGeneralizedLocomotion2022} and MorAL~\cite{luoMorALLearningMorphologically2024}, we parametrize the $c_i^\text{PD}$ as a function of the robot's total mass. Both polynomial mapping (as MorAL) and linear function (GenLoco) are considered. The latter additionally randomizes  $c_i^\text{PD}$ with a uniformly sampled value between $0.7$ and $1.1$.
    \item \textbf{Uniform randomization:} Following the \textit{uniform randomization} procedure in PAL~\cite{rytzReferenceFreePlatform2025}, all $c_i^\text{PD}\in\mathbb{R}^{2}$ are drawn uniformly within $\text{SR}=1$.
    \item \textbf{Interpolation with nominal sets:} Inspired by URMA~\cite{bohlingerOnePolicyRun2025}, we interpolate between a set of predefined nominal gains associated with different robots and apply a limited uniformly sampled noise, ie, $\text{SR}=\text{small}$. For robots between these reference values, interpolation is applied jointly on both the nominal gains and their 
    associated SR as a function of total mass. 
    \item \textbf{Performance-based adaptive particle filtering:} Following the computation of Equation~(\ref{eq:sr}) and~(\ref{eq:w_k_i_particleUp}), we update $n_\text{SR}, n_w,$ and $n_r$ configurations scaling SR of the joint PD gain parameters adaptively. 
\end{itemize}
The nominal PD gain values for MorAL, PAL,  URMA, ManyQuadrupeds, and GenLoco are indicated with an asterisk in Table~\ref{tab:pal_and_moral_velocityTracking} and as a red cross in Fig.~\ref{fig:results:pdgain}.

\subsubsection{Domain Randomization}
For successful sim-to-real transfer of policies trained in simulation dynamics, randomization is an important aspect~\cite{tanSimtoRealLearningAgile2018, xieDynamicsRandomizationRevisited2021}. Many morphology and dynamics parameters are randomized through the robot generation described in Section \ref{ss:robot_generation}. In any case  we additionally rescale for all $j$ in $c_{i_j}^\text{PD}\in\mathbb{R}^{24}$ by
$\widetilde{c}_{i_j}^\text{PD} = c_{i_j}^\text{PD}\cdot\text{clip}
 \left(1+ \epsilon, 0.95, 1.05
 \right)$
where $\epsilon\sim \mathcal{N}(0,1)$ is a standard normal variable scaling the PD gain, and $\text{clip}(\cdot)$ enforces lower and upper bounds. We introduce actuation command tracking delays and dynamics as in~\cite{rytzReferenceFreePlatform2025}.

\subsection{Training Setup - RL Algorithm}
\label{sec:methodology:training_setup}
We train the locomotion policy using Proximal Policy Optimization (PPO)~\cite{schulmanProximalPolicyOptimization2017}. The locomotion policy $\pi$ and the estimator policy are parameterized as a Multi-Layer Perceptron (MLP) with leaky-relu activation. The locomotion policy is modelled with hidden layers of size [512, 256, 128], and the estimator policy is modelled with layers of size [512, 256, 64]. We apply a penalty of $-0.25$ for early termination if a collision is detected as a self-collision or ground collision with a body other than a foot. The training time and hyperparameters were hand-tuned based off~\cite{rytzReferenceFreePlatform2025} and are provided in Table~\ref{tab:method:hyperparameters}. We trained the policies using 8 CPU cores (@4 GHz) and an NVIDIA RTX 4090 GPU. The hyperparameters used for the algorithm choice are described in Table~\ref{tab:method:hyperparameters} with $50000$ training steps taking approximately \SI{34}{\hour}. All training is done in simulation, using RaiSim~\cite{raisimtechArticulatedSystemsRaiSim}.
\begin{table}[h!]
    \centering
    \small
    \setlength{\tabcolsep}{3pt}
    \begin{tabular}{|l|c||l|c|}
        \hline
        \textbf{Parameter} & \textbf{Value} & \textbf{Parameter} & \textbf{Value} \\
        \hline
        Discount factor, $\gamma$ & 0.997 & Entropy coefficient & 0 \\
        Learning rate & adaptive & Value coefficient & 0.5 \\
        Batch size & 60750 & GAE & True \\
        Mini-batch size & 10 & GAE $\lambda$ & 0.95 \\
        Epochs & 6 & Steps per iteration & 135 \\
        Parallel envs, $n_{env}$ & 450 & PPO time/iteration & $\sim$\SI{2.2}{\second} \\
        \hline 
    \end{tabular}
    \caption{RL Training time and hyperparameters.}
    \label{tab:method:hyperparameters}
    \vspace{-5mm}
\end{table}

\section{Simulation and hardware validation}
\label{sec:results_and_discussion}

In this section, we compare the effectiveness of our controllers on the locomotion task for quadrupedal robots with different kinematic and dynamic parameters. 


\subsection{Performance Benchmark}
The comparative evaluation is performed among the following reference-free locomotion controllers for the flat terrain walking task. Unless noted otherwise, we compare the state-of-the-art uniform range sampling ($\text{SR}=1$) with the performance-based particle filter approach (SR adaptive) described in Section~\ref{ss:robotModelGen}.
\begin{enumerate}
    \item \textbf{Particle Filter, adaptive sampling range (Ours):} For the performance based $n_u=10\%\cdot n_p$ samples we adjust the SR following Equation~(\ref{eq:sr}). Additionally, we resample $n_r=10\%\cdot n_p$ replay buffer samples, and $n_w=10\%\cdot n_p$ particle filter weighted samples as per Equation~(\ref{eq:w_k_i_particleUp}). The PD gains are sampled on the available range clipped by SR.
    \item \textbf{PAL}~\cite{rytzReferenceFreePlatform2025}\textbf{:} We resample $n_r=15\%\cdot n_p$ with $\text{SR}=1$ and $n_r=\%15\cdot n_p$. The PD gains are sampled uniformly with $\text{SR}=1$.
    \item \textbf{GenLoco}~\cite{fengGenLocoGeneralizedLocomotion2022}\textbf{:} The PD gains are sampled using a linear function mapping robot weight to joint PD gains and additional uniform noise range. The original work only trains with a fixed set of robots, whereas we enable larger unseen parameter ranges; thus, we interpolate between nominal values to better cover robot mass values out of bounds.   
    \item \textbf{ManyQuadrupeds}~\cite{shafieeManyQuadrupedsLearningSingle2023}\textbf{:} Depending on the sampled total weight given a $\mathbf{c}_i$ we interpolate between the nominal PD gain values assigned to each reference robot model (similar to GenLoco).
    \item \textbf{MorAL}~\cite{luoMorALLearningMorphologically2024}\textbf{:} We apply the polynomial function mapping from total robot weight to scaled joint PD gains. 
    \item \textbf{URMA}~\cite{bohlingerOnePolicyRun2025}\textbf{:} We apply the same nominal PD gains with tight uniform noise ranges $\text{SR}=\text{small}$. The parameters are interpolated from the values given in the original work.
\end{enumerate}
Only our approach and PAL achieved stable locomotion on hardware, whereas MorAL and URMA maintained standing but failed to realize sustained walking (see supplementary video). GenLoco and ManyQuadrupeds were not evaluated on hardware, as simulation results showed reduced robustness compared to PAL and our method (see Fig.~\ref{fig:results:robustnessA1andANYmalC}) for ANYmal. 

In particular, their mass-dependent gain mappings required derivative gains approaching or exceeding validated hardware-safe limits (e.g., $K_d \geq 1.5$). Operating in this regime led to pronounced actuator impacts and audible mechanical knocking during preliminary trials (see supplementary video), which are indicative of aggressive controller dynamics and elevated actuator loading. Such behavior increases the risk of actuator wear, structural damage, or unstable oscillations. Given that these approaches neither demonstrated competitive robustness in simulation nor operated within established safe gain ranges for the platform, they were excluded from hardware evaluation to prevent potential damage to the system.

\subsection{Sim-to-Sim Robustness}
\begin{figure}
    \centering
    \includegraphics[width=0.9\columnwidth]{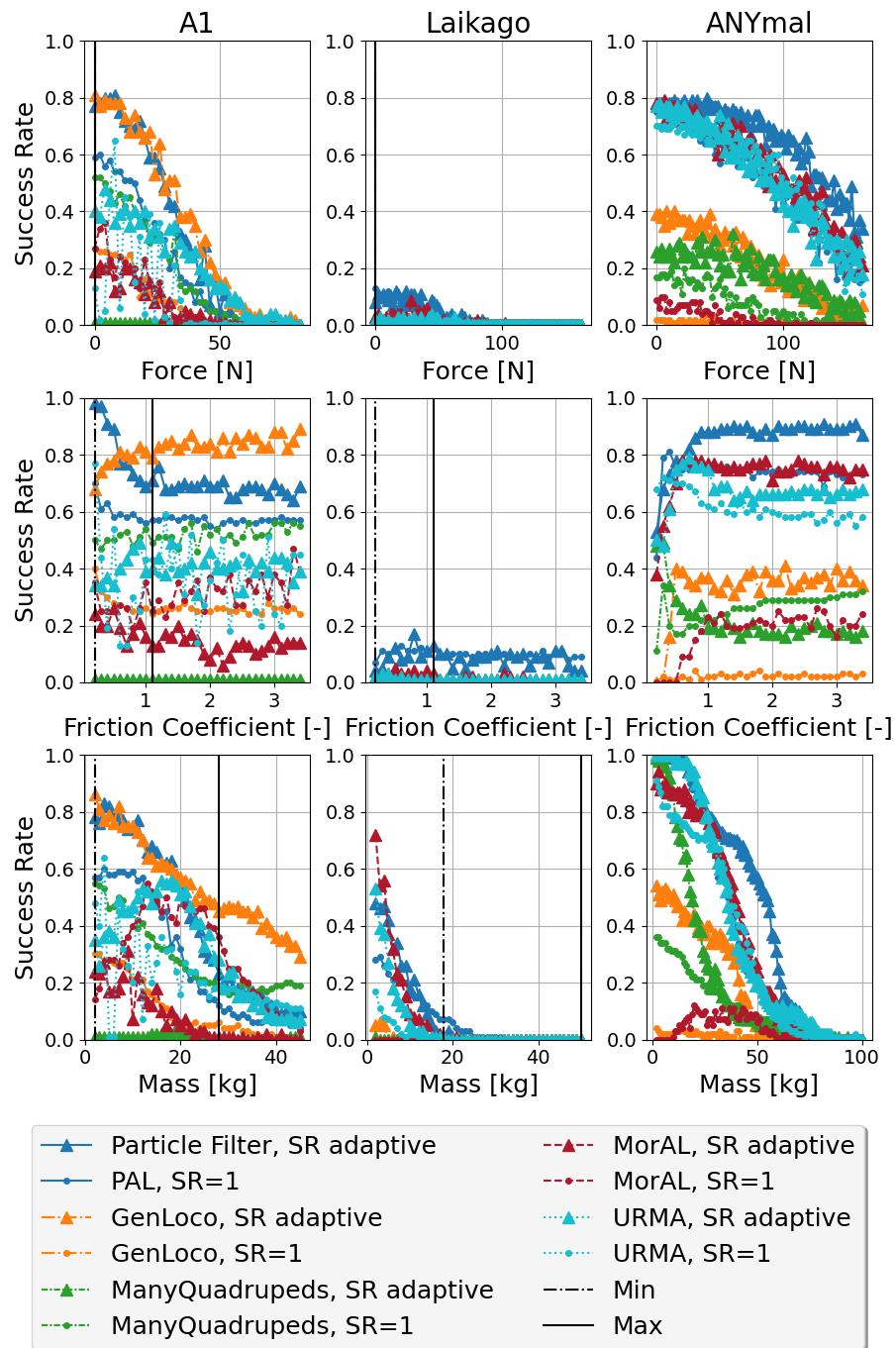}
    \caption{Success rates for various parameter sampling strategies for different perturbations and dynamics parameters for the A1 (left), Laikago (middle), and ANYmal (right) quadrupeds. During random walking commands, we measure $\text{SR}^*$ over base perturbations in the horizontal plane, ground friction coefficient, and changes in base mass, and show in black the minimum and maximum training ranges of the disturbance parameters.}
    \vspace{-0.6cm}
    \label{fig:results:robustnessA1andANYmalC}
\end{figure}

We use the definition of success rate ($\text{SR}^*$) as a performance metric for robustness as: $\text{SR}^* = 1 - \frac{N_e}{N_T}.$ With $N_e$ referring to the number of rollouts that terminated early due to a prohibited collision and $N_T$ being the total number of rollouts. Using $N_T=100$, we randomize the base linear velocity command for \SI{4}{\second} before re-spawning while applying the same early termination criteria defined in Section~\ref{ss:robot_generation}. To assess cross-morphology generalization, we evaluate policies on three quadrupeds of increasing scale: A1 (simulation-tuned PD gains $(35,0.5)$), Laikago (simulation-tuned PD gains $(32,0.6)$), and ANYmal (hardware-tuned PD gains $(85,0.5)$). Importantly, Laikago is not used as a reference model during training and serves as an \emph{unseen} morphology to assess zero-shot sim-to-sim transfer. We test three parameter changes, measuring each time the $\text{SR}^*$: horizontal force perturbation onto the base, changes in foot contact friction, and base mass. The resulting robustness curves are depicted in Fig.~\ref{fig:results:robustnessA1andANYmalC}. 

Overall, the results reveal several consistent trends across robots and sampling strategies. First, adaptive sampling of the configuration space improves robustness across all evaluated quadrupeds compared to fixed uniform sampling with $\text{SR}=1$, except for MorAL applied to the A1 quadruped. Second, while GenLoco performs competitively on A1, this strategy does not transfer to the larger ANYmal platform. This suggests that gain mappings tuned to imitation-based trajectories or central pattern generators do not scale reliably across substantially different robot morphologies. Third, zero-shot transfer to out-of-distribution morphologies remains challenging. As illustrated by the Laikago results, none of the evaluated approaches achieve consistently high $\text{SR}^*$ across the tested perturbations, indicating that substantial morphological shifts beyond the training distribution degrade locomotion performance. Finally, URMA demonstrates strong robustness among the state-of-the-art approaches, motivating our curriculum-based design. However, both PAL with $\text{SR}=1$ and our adaptive particle filter consistently achieve equal or slightly higher robustness across A1 and ANYmal.

\begin{figure}[h]
    \centering
    \includegraphics[width=\columnwidth]{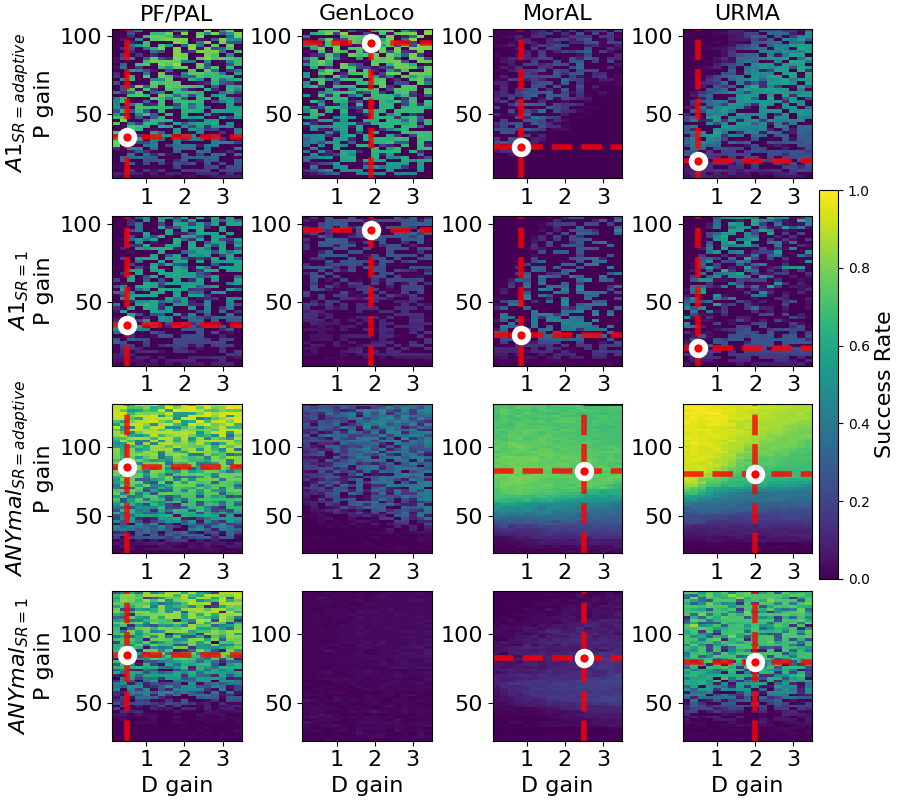}
    \caption{The success rate $\text{SR}^*$ is measured over a range of PD gain combinations, with yellow indicating good locomotion capabilities and dark blue indicating low to no stable performance. The red-white dot indicates the nominal PD gains as per the original implementation. The tests were run in RaiSim for the A1 (top two rows) and ANYmal (bottom two rows). The different $\mathbf{c}_i$ sampling strategies go from left to right as Particle Filter (SR adaptive) or PAL (uniform sampling, i.e. $\text{SR}=1$), GenLoco ($\text{SR}=1$ or adaptive), MorAL ($\text{SR}=1$ or adaptive), and URMA ($\text{SR}=$small or adaptive). We removed ManyQuadrupeds as it performed worse than GenLoco in the robustness test shown in Fig.~\ref{fig:results:robustnessA1andANYmalC}. In red, we show the nominal gains for the respective sampling type and robot.}
    \vspace{-0.5cm}
    \label{fig:results:pdgain}
\end{figure}

Fig.~\ref{fig:results:pdgain} provides further insight into the influence of PD gain sampling on robustness. The particle-filter-based adaptive strategy consistently exhibits the largest contiguous regions of high $\text{SR}^*$ (yellow areas), indicating stable locomotion across a wider range of PD gain combinations. This broader coverage is particularly pronounced for the
larger ANYmal platform and aligns with the improved hardware robustness reported in Section~\ref{ss:results:sim_2_real_transfer} and demonstrated in the supplementary video.

In contrast to Particle Filter, PAL, MorAL, or URMA, GenLoco shows an almost absent region of stable performance for ANYmal within hardware-safe PD gain ranges for deployment. Since we neither observed sufficient $\text{SR}^*$ coverage in simulation nor found evidence in the prior literature of safely deploying these high gains on hardware, GenLoco (and ManyQuadrupeds) was not evaluated on ANYmal. URMA demonstrates comparatively broad stability regions around nominal gains, confirming that injecting prior knowledge via sampling around reference values is beneficial. However, both PAL with $\text{SR}=1$ and our adaptive particle filter achieve equal or larger stable regions across A1 and ANYmal. Visually, the particle filter approach yields the most extensive coverage of high-performance PD gains, supporting the conclusion that performance-adaptive sampling enhances robustness for zero-shot deployment.

\subsection{Sim-to-Real Transfer}
\label{ss:results:sim_2_real_transfer}

In our hardware experiments, we transferred the trained dynamics encoding and locomotion policy networks to the ANYmal quadruped. 

\subsubsection{Hardware Velocity Command Tracking}
We evaluate the effectiveness of the learned controllers directly on the ANYmal platform. 
Table~\ref{tab:pal_and_moral_velocityTracking} reports the Root Mean Square Error (RMSE) between commanded and estimated base velocities in the heading ($X$), lateral ($Y$), and yaw ($\theta$) directions, together with a deployment indicator. The deployment symbols denote hardware performance as follows: \textbf{$\checkmark$} indicates safe and reliable hardware execution, \textbf{$\diamond$} denotes partial deployment (the controller executed but exhibited instability or degraded tracking performance), and 
\textbf{$\times$} indicates unsafe deployment, where the robot failed to walk (e.g., collapse, progressive and unsafe base sagging, or refusal to locomote at all) and therefore did not yield valid RMSE measurements. The superscript 13 denotes the results for locomotion performance with an additional payload of \SI{13}{\kilo\gram} mounted on top of the quadruped.

\begin{table}[h!]
    \centering
    \renewcommand{\arraystretch}{1.15}
    \begin{tabular}{|l@{\hspace{2pt}}|c@{\hspace{2pt}}|c@{\hspace{2pt}}|c@{\hspace{2pt}}|c@{\hspace{2pt}}|c@{\hspace{2pt}}|c@{\hspace{2pt}}|}
        \hline
        \textbf{Sampling}          & \textbf{SR} & \textbf{Safe?}   & \textbf{P:D gains} & $\mathbf{X}$ & $\mathbf{Y}$ & $\mathbf{\theta}$ \\
        \hline
        
        Particle Filt & adap & \textbf{\checkmark} & 85.0:0.5 & 0.0954 & 0.0706 & 0.2278 \\
        Particle Filt & adap & \textbf{\checkmark} & 85.0:1.0 & 0.1100 & 0.0764 & 0.2530 \\

        $\text{Particle Filt}^{13}$ & adap & \textbf{\checkmark} & 85.0:0.5 & 0.1041 & 0.0698 & 0.2922 \\
        \hline
        
        PAL         & 1     & \textbf{\checkmark}  & 85.0:0.5 & 0.0947 & 0.0770 & 0.2504 \\
        PAL         & 1       & \textbf{\checkmark} & 85.0:1.0 & 0.1082 & 0.0878 & 0.2955 \\

        $\text{PAL}^{13}$         & 1     & \textbf{\checkmark}  & 85.0:0.5 & 0.1070 & 0.0772 & 0.2199 \\
        \hline
        
        GenLoco         & 1/adap      & \textbf{$\times$} & $400:8.^*$ & - & - & - \\
        \hline
        
        ManyQuad  & 1/adap       & \textbf{$\times$} & $430:20.7^*$ & - & - & - \\
        \hline        
        
        MorAL           & 1       & \textbf{$\times$} & $82:5^*$ & - & - & - \\
        MorAL           & adap & \textbf{$\diamond$} & 85.0:1.0 & 0.1864 & 0.1227 & 0.3384 \\
        \hline

        URMA       & 1        & \textbf{$\times$} & $80:2^*$ & - & - & - \\
        URMA       & adap & \textbf{$\diamond$} & 85.0:1.0 & 0.1179 & 0.0881 & 0.1540 \\
        \hline
        
    \end{tabular}
    \caption{Walking test: velocity command tracking error comparison with RMSE of heading (X), lateral (Y), and turn ($\theta$) commands on the hardware ANYmal.}
    \label{tab:pal_and_moral_velocityTracking}
    \vspace{-6mm}
\end{table}

The results indicate that the proposed Particle Filter sampling strategy achieves tracking performance comparable to the PAL baseline, with similar RMSE values across heading, lateral, and yaw velocities. While neither method consistently outperforms the other across all metrics, both enable reliable and stable hardware deployment. Particle Filter and PAL sampling strategies achieve consistent and safe real-world locomotion, while adaptive variants of URMA and MorAL exhibit degraded but executable behavior, differing from the full sampling ranges reported in their original formulation with SR=1 in this work. All remaining configurations were unsafe for hardware deployment following the simulation evaluation.


\subsubsection{Hardware Base Velocity Estimator}
The velocity estimate is another measure of the quality of the controllers and robot IDs chosen. We compute the RMSE of the trained MLP estimator network with the onboard manufacturer ANYmal state estimator in Table \ref{tab:pal_and_moral_baseEstimator}.

\begin{table}[h!]
    \centering
    \renewcommand{\arraystretch}{1.15}
    \begin{tabular}{|l@{\hspace{2pt}}|c@{\hspace{2pt}}|c@{\hspace{2pt}}|c@{\hspace{2pt}}|c@{\hspace{2pt}}|c@{\hspace{2pt}}|c@{\hspace{2pt}}|}
    \hline
        \textbf{Sampling}          & \textbf{SR} & \textbf{Safe?} & \textbf{P:D gains} & $\mathbf{X}$ & $\mathbf{Y}$ & $\mathbf{Z}$ \\
        \hline

        Particle Filt           & adap & \textbf{\checkmark} & 85.0:0.5 & 0.1697 & 0.1595 & 0.0881 \\
        Particle Filt           & adap & \textbf{\checkmark} & 85.0:1.0 & 0.1771 & 0.1053 & 0.0824 \\

        $\text{Particle Filt}^{13}$       & adap & \textbf{\checkmark} & 85.0:0.5 & 0.1477 & 0.1733 & 0.0872 \\
        \hline
        
        PAL                 & 1       & \textbf{\checkmark} & 85.0:0.5 & 0.1339 & 0.1181 & 0.0797 \\
        PAL       & 1       & \textbf{\checkmark} & 85.0:1.0 & 0.1384 & 0.1148 & 0.0601 \\

        $\text{PAL}^{13}$   & 1       & \textbf{\checkmark} & 85.0:0.5 & 0.1030 & 0.1322 & 0.0734 \\
        \hline
                
        GenLoco         & 1/adap      & \textbf{$\times$} & $400:8.^*$ & - & - & - \\
        \hline

        ManyQuad  & 1/adap      & \textbf{$\times$} & $430:20.7^*$ & - & - & - \\
        \hline
        
        MorAL           & 1       & \textbf{$\times$} & $82:5^*$ & - & - & - \\
        MorAL           & adap & \textbf{$\diamond$} & 85.0:1.0 & 0.0942 & 0.0797 & 0.1080 \\
        \hline

        URMA       & 1       & \textbf{$\times$} & $80:2^*$ & - & - & - \\
        URMA       & adap & \textbf{$\diamond$} & 85.0:1.0 & 0.0863 & 0.0701 & 0.0762 \\
        \hline
    \end{tabular}
    \caption{Walking test: base velocity estimate RMSE of heading (X), lateral (Y), and height (Z) estimate with onboard state estimator base twist for hardware ANYmal.}
    \vspace{-0.3cm}
    \label{tab:pal_and_moral_baseEstimator}
\end{table}

Either sampling approach, PAL and Particle Filter, is robust enough for zero-shot hardware deployment. PAL sampling yields better velocity estimation, which we attribute to the estimator network's early exposure to the full range of possible $\mathbf{c}_i$ parameters from the start of the training phase. These results show that larger PD gain sampling ranges during training enable zero-shot \textit{universal} locomotion.

\section{Conclusion}
\label{sec:conclusion_and_future_work}

In this paper, we investigated the role of configuration parameter sampling for \textit{universal} quadrupedal locomotion. We showed that fixed uniform sampling and mass-dependent PD gain mappings can lead to reduced robustness or unsafe hardware deployment when scaling across morphologies and controller gains, particularly for larger platforms such as ANYmal. To address these limitations, we introduced a performance-based adaptive curriculum that reshapes morphology and PD gain distributions during training via sampling range adaptation and particle-filter-based resampling. Extensive simulation experiments on A1, ANYmal, and an unseen Laikago morphology demonstrate that adaptive sampling expands stable operating regions under disturbance and gain variations. Hardware experiments on ANYmal further confirm reliable zero-shot deployment, including locomotion under an additional +25\% payload mass exceeding nominal manufacturer specifications, while respecting safety constraints.

Although zero-shot transfer to strongly out-of-distribution morphologies remains challenging, performance-adaptive sampling substantially improves robustness within a broad and practically relevant morphology range. Future work will focus on integrating perceptive components for terrain-aware universal locomotion and validating the approach across a wider range of quadrupeds and hardware scales.

\bibliographystyle{style/IEEEtran}
\bibliography{references.bib}

\end{document}